\def\year{2016}\relax
\documentclass[letterpaper]{article}
\usepackage{aaai16}
\usepackage{times}
\usepackage{helvet}
\usepackage{courier}
\usepackage[utf8]{inputenc}
\usepackage{booktabs}
\usepackage{array}
\usepackage{ragged2e}
\usepackage{graphicx}
\usepackage{multirow}
\usepackage{multicol}
\newcolumntype{P}[1]{>{\RaggedRight\hspace{0pt}}p{#1}}

\newcommand{\ie}{i.e.,\ }
\def\etal{\emph{et al}. }
\newcommand{\qt}[1]{``#1"}
\newcommand{\Tab}[1]{Table~\ref{#1}}

\newcommand{\Fig}[1]{Figure~\ref{#1}}
\newcommand*{\postagaccuracyturkish}{96.85 }
\newcommand*{\stemmingaccuracyturkish}{97.59 }
\newcommand*{\accuracyturkish}{84.12 }
\newcommand*{\postagaccuracygerman}{98.35 }
\newcommand*{\stemmingaccuracygerman}{95.95 }
\newcommand*{\accuracygerman}{88.35 }
\newcommand*{\postagaccuracyfrench}{98.47 }
\newcommand*{\stemmingaccuracyfrench}{99.52 }
\newcommand*{\accuracyfrench}{93.78 }
\newcommand*{\accuracyWithRootEmb}{85.18 }

\begin{document}
\title{A Morphology-aware Network for Morphological Disambiguation}
\author{
	Eray Yildiz \And
	Caglar Tirkaz \And
	H. Bahadir Sahin 	\\\\
	*Huawei Turkey Research and Development Center, Umraniye, Istanbul, Turkey
	\\{\tt \{eray.yildiz, mustafa.tolga.eren\}@huawei.com}\\ {\tt \{caglartirkaz, hbahadirsahin, osonmez\}@gmail.com} \And	
	Mustafa Tolga Eren \And
	Ozan Sonmez	
}
\maketitle

\begin{abstract}
\begin{quote}
Agglutinative languages such as Turkish, Finnish and Hungarian require morphological disambiguation before further processing due to the complex morphology of words. A morphological disambiguator is used to select the correct morphological analysis of a word. 
Morphological disambiguation is important because it generally is one of the first steps of natural language processing and its performance affects subsequent analyses. 
In this paper, we propose a system that uses deep learning techniques for morphological disambiguation. 
Many of the state-of-the-art results in computer vision, speech recognition and natural language processing have been obtained through deep learning models. 
However, applying deep learning techniques to morphologically rich languages is not well studied. 
In this work, while we focus on Turkish morphological disambiguation we also present results for French and German in order to show that the proposed architecture achieves high accuracy with no language-specific feature engineering or additional resource. 
In the experiments, we achieve \accuracyturkish, \accuracygerman and \accuracyfrench morphological disambiguation accuracy among the ambiguous words for Turkish, German and French respectively.
\end{quote}
\end{abstract}

\section{Introduction}

Morphological analysis is generally achieved through the use of finite state transducers (FST) \cite{kaplan1981phonological,koskenniemi1984general,beesley2003finite,Oflazer:1993:TDT:976744.976810}. 
During morphological analysis, the surface form of the word is given as input and an FST is used to output possible morphological analyses of the input word. 
A morphological analysis contains a root and a set of tags so called morphemes, minimal units of meaning in a language \cite{Oflazer:1993:TDT:976744.976810}. 

A morphological disambiguator is used to select the correct analysis among the possible analyses of a word using the context that the word appears in. 
The output of morphological disambiguation contains syntactic and semantic information about a word such as its POS tag, tense, polarity and it being accusative, possessive or genitive. 
This information is vital for some NLP tasks such as dependency parsing and semantic role labeling whereas it can be utilized in other NLP tasks such as topic modeling, named entity recognition and machine translation. 

While morphological disambiguation is important for natural language processing in any language, it is vital in morphologically rich languages. 
We specifically focus on Turkish which is an important language spoken by over 70 million people and has a complex morphology that allows construction of thousands of word forms from each root through inflectional and derivational suffixation \cite{Hakkani-Tur:2000:SMD:990820.990862}. 
For instance, \textit{yürü (walk)}, \textit{yürüdüm (I walked)}, \textit{yürüyeceksiniz (you will walk)}, \textit{yürüttüler (they made somebody walk)}, \textit{yürüyünce (When (he/she/it) walks)} and \textit{yürüyecektiler (They were going to walk)} are some of the possible word formations of a Turkish verb root \textit{yürü}. In all the examples \qt{yürü} is the root of the words whereas the suffixes are used to change meaning. 

Morphological analysis of a word might produce more than one analysis since there might be multiple interpretations of a single word. Consider the example given in \Tab{tab:wordformations2} where the Turkish word \qt{dolar} is analyzed. The output of the morphological analyzer for this word contains four possible analyses. The reason for that is each of the words \qt{dolar}, \qt{dola}, \qt{dol} and \qt{do} can be used as roots and at the same time \qt{r}, \qt{ar} and \qt{lar} are all valid suffixes in Turkish. Thus, all four of the analyses are valid that lead to quite different meanings. 

\begin{table}[!htb]
	\caption{Morphological analyses of the Turkish word \qt{dolar}}	
	\centering
	\begin{tabular}{P{0.65\linewidth}P{0.25\linewidth}}
	\toprule
		Morphological & English \\
		Analysis & Translation \\
		\midrule
		dolar +Noun +3sg +Pnon +Nominative & dollar\\
		dola +Verb +Positive +Aorist +3sg &he/she wraps\\
		dol +Verb +Positive +Aorist +3sg&it fills\\	
		do +Noun +3pl +Pnon +Nominative &Multiple C (musical note)\\
		\bottomrule
	\end{tabular}
	\label{tab:wordformations2}	
\end{table}	

Another reason for multiple morphological analyses is due to the fact that a morpheme might change meanining depending on the context of the word. 
Consider the example given in \Tab{tab:wordformations3}. 
In the first row \qt{evi} is used in the accusative case whereas in the second row it is used as a possessive noun. 
The word \qt{evi} has two morphological analyses sharing the same root. Thus, the only difference in the analyses is at the suffix of the word. 
The suffix \qt{–i} in the first sentence transforms the word into  \qt{accusative marker}, while its interpretation is \qt{third person possessive} in the second one. In addition, some root words might have multiple meanings. For instance, the Turkish word \qt{yüz} could be interpreted as a noun (face), a verb (swim) or a number (hundred) depending on its context. 

\begin{table}[!ht]
	\caption{Multiple interpretations of the Turkish word \qt{evi}}	
	\centering
	\begin{tabular}{P{0.55\linewidth}P{0.34\linewidth}}
		\toprule 
		Turkish sentence and its translation & Morphological Analysis \\
		\midrule
		\textbf{Evi} bulabildiniz mi?
		-- Did you find the house? & ev +Noun +3sg +Pnon +Accusative\\ 
		\textbf{Evi} gerçekten güzelmiş. -- His/Her house is really beautiful. & ev +Noun +3sg +P3sg +Nominative\\ 
		\bottomrule
	\end{tabular}
	\label{tab:wordformations3}	 
\end{table}

As we noted before, morphological disambiguation is important for NLP in most of the languages. 
For instance, although German and French do not have a morphology as rich as Turkish, NLP in these languages can still benefit from morphological disambiguation. Higher accuracies in NLP tasks such as POS tagging and dependency parsing can be obtained if the morphology of the words are taken into account \cite{sennrich2009new}, \cite{candito2010benchmarking}. 
We apply our general purpose morphological disambiguation method to these languages and show that high accuracies can be obtained for POS tagging and lemmatization. 
Possible word formations and morphological analyses of the German word \qt{Haus} and the French word \qt{savoir} are given in \Tab{tab:wordformations4} and \Tab{tab:wordformations5} respectively.

\begin{table}[!ht]
	\centering
	\caption{Possible word formations and morphological analyses of the German word \qt{haus}}
	\label{tab:wordformations4}
	\begin{tabular}{P{0.2\linewidth}P{0.7\linewidth}}
		\toprule
		Word & Morphological Analyses          \\ \midrule
		haus         & Noun Neuter Nominative Singular \\
		              & Noun Neuter Dative Singular     \\
		              & Noun Neuter Accusative Singular \\
		häuser       & Noun Neuter Accusative Plural   \\
		              & Noun Neuter Nominative Plural   \\
		              & Noun Neuter Genitive Plural     \\
		häusern      & Noun Neuter Dative Plural       \\
		hause      & Noun Neuter Dative Singular     \\
		hauses       & Noun Neuter Genitive Singular   \\ \bottomrule
	\end{tabular}
\end{table}

\begin{table}[!ht]
	\centering
	\caption{Possible word formations and morphological analyses of the French word \qt{savoir}}
	\label{tab:wordformations5}
	\begin{tabular}{P{0.2\linewidth}P{0.7\linewidth}}
	\toprule
	Word & Morphological Analyses                  \\ \midrule
	savoir      & Noun Masculine Singular                 \\
	            & Verb Infinitive                         \\
	sais         & Verb Present SecondPerson Singular      \\
	           & Verb Present FirstPerson Singular       \\
	savons       & Verb Present FirstPerson Plural         \\
	savaient    & Verb Imperfect ThirdPerson Plural       \\
	saches      & Verb Subjunctive SecondPerson Singular  \\
	sachant      & Verb Present Participle                 \\
	su         & Verb Past Participle Masculine Singular \\ \bottomrule
	\end{tabular}
\end{table}

There are various approaches proposed for morphological disambiguation based on lexical rules or statistical models. Rule based methods apply hand-crafted rules in order to select the correct morphological analyses or eliminate incorrect ones \cite{Oflazer:1994:TMD:974358.974391,W96-0207,Daybelge07arule-based}. Yüret and Türe (\citeyear{Yuret:2006:LMD:1220835.1220877}) proposed a decision list learning algorithm for extraction of Turkish morphological disambiguation rules from disambiguated training data. Tür \etal (\citeyear{Hakkani-Tur:2000:SMD:990820.990862}) developed a statistical model which scores the probability of each analysis using trigram models of the tags and roots. Sak \etal (\citeyear{sak-et-al-cicling-07}) applied a multilayer perceptron algorithm which uses n-grams of the roots and tags as features. 
A CRF based disambiguation model was proposed by Razieh \etal \citeyear{DBLP:conf/rocling/EhsaniAEA12}. 
Finally, hybrid models which combine statistical and rule based approaches are also proposed \cite{W96-0207,DBLP:conf/ijcnlp/KutluC13}.

We propose a deep neural architecture followed by the Viterbi algorithm for morphological disambiguation of words in a sentence. 
In this paper we focus on Turkish as an example even though the proposed model can be utilized in all morphologically rich languages. 
We test our approach in German and French in order to prove that the proposed method is able to work well for other languages as well. 
The network architecture in this paper is designed to produce a classification score for a sequence of \textit{n}-words. It consists of two layers and a softmax layer. 
The first layer of the model builds a representation for each word using root embeddings and some syntactic and semantic features. 
The second layer takes as input the learned word representations and incorporates contextual information. 
A softmax layer uses the output of the second layer to produce a classification score. 
We use the neural network to produce a score for each \textit{n} length sequence in a given sentence. 
We then employ the Viterbi algorithm to produce the morphological disambiguation for each word in the sentence by finding the most probable sequence using the output of the softmax layer.

\section{Related Works}	
\label{sec:Related Works}
In a natural language processing pipeline morphological disambiguation can be considered at the same level as POS tagging. 
In order to perform POS tagging in English, various approaches such as rule-based models \cite{Brill:1992:SRP:974499.974526}, statistical models \cite{Brill:1995:TEL:218355.218367}, maximum entropy models \cite{Ratnaparkhi1997}, HMMs \cite{Cutting:1992:PPT:974499.974523}, CRFs \cite{lafferty2001conditional} and decision trees \cite{Schmid94probabilisticpart-of-speech} are proposed. 
However, morphological disambiguation is a much harder problem in general due to the fact that it requires the classification of both roots, suffixes and the corresponding labels. 
Moreover, compared to an agglutinative language such as Turkish, English words can take on a limited number of word forms and part-of-speech tags. 
Yüret and Türe (\citeyear{Yuret:2006:LMD:1220835.1220877}) observe that more than ten thousand tag types exists in a corpus comprised of a million Turkish words. 
Thus, due to the high number of possible tags and the number of possible analyses in languages with productive morphology, morphological disambiguation is quite different from part-of-speech tagging in English.

The previous work on morphological disambiguation in morphologically rich languages can be summarized into three categories: rule based, statistical and hybrid approaches. 
In the rule-based approaches a large number of hand crafted rules are used to select the correct morphological analyses or to eliminate incorrect ones \cite{Karlsson:1995:CGL:546590,Oflazer:1994:TMD:974358.974391,W96-0207,Daybelge07arule-based}. 
Statistical approaches generally utilize the statistics of root and tag sequences for selection of the best roots and tags. A statistical Turkish morphological disambiguation model which scores the probability of each tag by considering the statistics over the derivational boundaries and roots using trigrams has been proposed by Tür \etal (\citeyear{Hakkani-Tur:2000:SMD:990820.990862}). They test their model on a manually disambiguated test data consisting of 2763 words and obtain 93.5\% accuracy in morphological disambiguation (including non-ambiguous words). A similar morphology-aware nonparametric Bayesian model is proposed in \cite{chahuneau2013knowledge}. They integrate their generative model to NLP applications such as language modeling, word alignment and morphological disambiguation and obtain state-of-the-art results for Russian morphological disambiguation. 
Yüret and Türe (\citeyear{Yuret:2006:LMD:1220835.1220877}) extract Turkish morphological disambiguation rules using a decision list learner, Greedy Prepend Algorithm (GPA), and they achieve 95.8\% accuracy on manually disambiguated data consisting of around 1K words. 
Megyesi (\citeyear{Megyesi99improvingbrill's}) adapt a transformation based syntactic rule learner \cite{Brill:1995:TEL:218355.218367} for Hungarian and Hajic (\citeyear{hajic1998czech}) extend his work for Czech and five other languages. 
Sak \etal (\citeyear{sak-et-al-cicling-07}) apply a multilayer perceptron algorithm using a set of 23 features including tri-gram and bi-gram statistics of morphological tags and roots. They obtain 96.8\% accuracy on test data consisting of 2.5K words. 
Ehsani \etal (\citeyear{DBLP:conf/rocling/EhsaniAEA12}) apply conditional random fields (CRFs) using several features derived from morphological and syntactic properties of words and achieve 96.8\% accuracy. 
Görgün and Yildiz (\citeyear{DBLP:conf/iscis/GorgunY11}) use a J48 decision tree and achieve 95.6\% accuracy. 
There are also several studies that combine statistical and rule based approaches such as \cite{ezeiza1998combining,W96-0207,DBLP:conf/ijcnlp/KutluC13,orosz2013purepos}. 

Although deep learning techniques have been successfully used in various NLP tasks in English\cite{Collobert:2008:UAN:1390156.1390177,Collobert:2011:NLP:1953048.2078186,icml2014c2_le14,pennington-socher-manning:2014:EMNLP2014,DBLP:conf/conll/LuongSM13,Socher:2012:SCT:2390948.2391084}, this study is unique in that we create a deep learning architecture specifically suited for handling morphologically rich languages. 
One similar work to ours is the recent work of Luong \etal (\citeyear{DBLP:conf/conll/LuongSM13}) who introduce morphological RNNs to create word representations through composition of morphemes. 
However, they present their results only for English which is not morphologically as rich as languages such as Turkish or Finnish. 
There are also recent works that suggest integrating morphological knowledge into distributed word representations such as \cite{cotterellmorphological} and \cite{cui2015knet}. 
Cotterell and Schütze (\citeyear{cotterellmorphological}) extend log-bilinear model (an instance of language models that make the Markov assumption as n-gram language models) in order to jointly predict the next morphological tag along with the next word, encouraging the resulting embeddings to encode morphology.
On the other hand, \cite{cui2015knet} propose a method for learning embeddings which is a modified version of skip-gram algorithm \cite{mikolov2013distributed} that benefits from morphological knowledge when predicting the target word. 
Using morphology-based word representations improves the performance for different NLP tasks such as word similarity and statistical machine translation according to empirical evaluation of Botha and Phil (\citeyear{botha2014compositional}).

Our work uses a convolutional architecture and handles any number of morphological features in order to build word representations while performing disambiguation at the same time. 

\section{Method}
\label{sec:Method}

\begin{figure*}[!ht]
	\centering
	\includegraphics{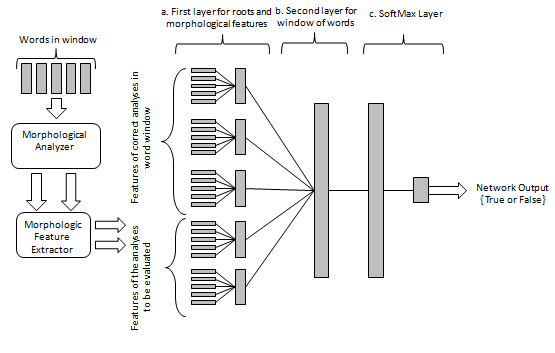}
	\caption{Structure of morphology-aware model. 
		\textbf{Layer (a)} construct word vectors using the morphological features of the input word.
		\textbf{Layer (b)} allows the model to utilize contextual information by considering a window of words. 
		\textbf{Layer (c)} is the softmax layer that makes a binary decision of whether or not the current disambiguation result is correct. 
	}
	\label{fig:architecture}
\end{figure*}

In this work we propose an architecture with the ability to represent morphologically rich words and model spatial dependencies among word vectors. A softmax layer that is trained on top of the layers is used to predict the likelihood of a window of words. Finally, the Viterbi algorithm is used on the outputs of the softmax layer in order to find the optimal disambiguation of the words in a sentence. 
We also show how unsupervised pre-trainining can be used to improve the performance of the designed system and achieve the state-of-the-art accuracy for Turkish morphological disambiguation. 

The input to our model is a sentence where each word in the sentence needs to be disambiguated. 
We first tokenize the sentences and then use morphological analyzers to find possible analyses of each word in the sentence. 
HFST tool \cite{linden2009hfst} is used to perform morphological analysis in German and French whereas \cite{Oflazer:1993:TDT:976744.976810} is used for Turkish morphological analysis. 
NLP systems that use deep learning generally employ word embeddings in order to represent each word in a dictionary.  Word embeddings are dense low dimensional real-valued vectors with each dimension corresponding to a latent feature of the word \cite{Turian:2010:WRS:1858681.1858721}. 
In a morphologically rich language, representing words in surface form might not be a good idea since lots of surface form words can be derived from a single root. 
Thus, in our design, each word in surface form is represented with a root and a set of morphological features where each root and feature has individual embeddings that are learned during training. 
Root and morphological feature embeddings can have varying lengths and through their concatenation surface form words are represented with fixed length embeddings. 

Our architecture is illustrated in \Fig{fig:architecture} where individual layers are marked with \textbf{(a)}, \textbf{(b)} and \textbf{(c)}. 
The first layer \textbf{(a)} takes the root and morphological features of a single word as input and propagates to the next layer. 
The second layer, \textbf{(b)}, takes a window of \textit{n} words as input and propagates to the softmax layer, \textbf{(c)}. 
The non-linearity in both the first and the second layers are provided through the use of \textit{tanh} as the transfer function. 
The softmax layer is responsible for deciding the likelihood of the current morphological analysis of the words, \ie a binary decision is produced with the expected result of 1 if the analysis is correct, 0 otherwise.

\begin{table*}[!t]
	\tiny
	\centering
	\caption{The morphological (morphosyntactic and morphosemantic) features we used to represent each word}
	\label{tab:features}
	\begin{tabular}{P{0.08\linewidth}P{0.04\linewidth}P{0.06\linewidth}P{0.06\linewidth}P{0.07\linewidth}P{0.06\linewidth}P{0.06\linewidth}P{0.07\linewidth}P{0.06\linewidth}P{0.06\linewidth}P{0.05\linewidth}P{0.05\linewidth}P{0.04\linewidth}}
		\toprule
		& \multicolumn{10}{c}{Morphosyntactic and Morphosemantic Features}                                                                                                                         \\ \cmidrule(l){2-11} 
		\newline  Language&  Root &  Main POSTag &  Minor POSTag & Person Agreement &  Plurality & Gender & Possesive Agreement & Case Marker & Polarity & Tense \\ \midrule
		Turkish  & +          & +                 & +                  & +                       & +               & -            & +                          & +                 & +              & +           \\
		German   & +          & +                 & +                  & +                       & +               & +            & -                          & +                 & -              & +           \\
		French   & +          & +                 & -                  & +                       & +               & +            & -                          & -                 & -              & +           \\ \bottomrule
	\end{tabular}
\end{table*}

We train our network with the possible sequences of morphological analyses in the training data. 
For each sentence, and for each word, we select the \textit{n-2} words preceding the word and their groundtruth annotations along with the possible annotations of the last two words. 
We also add \textit{n-1} out of sentence tokens at the beginning of each sentence so that all words in the sentence are included in the training data. 
We label the sequences containing the correct morphological analysis as positive whereas the remaining sequences are labeled as negative. 
This way the model is trained to predict the correct annotation for the last two words in a sequence given that the first \textit{n-2} words have correct annotations. 
Training is performed with stochastic gradient descent and AdaGrad  \cite{duchi2011adaptive} as the optimization algorithms. 
At inference time, given a sentence containing words to disambiguate, we use the network to make predictions for window of words in the sentence and then use the Viterbi algorithm to select the best morphological analysis for each word. 

Unsupervised pre-training of word embeddings have been employed in various NLP tasks, and their usage have improved recognition accuracies \cite{Collobert:2011:NLP:1953048.2078186,Turian:2010:WRS:1858681.1858721}. 
In order to improve the performance of our disambiguation system we also use unsupervised methods to pre-train root embeddings of words. 
We created a corpus comprised of 1 billion Turkish words that we collected from various sources, such as e-books and web sites. 
Although our corpus is rather small compared to English corpora, it is the largest text corpus in Turkish that we know of. 
After we trained the supervised disambiguation system as described above, we disambiguated each word in the corpus and extracted the roots of words. 
Next, we built representations for root forms of the words using the unsupervised skip-gram algorithm \cite{mikolov2013distributed}. 
After obtaining the pre-trained root vectors, we retrained our disambiguation system with pre-trained root embeddings. 
This technique allowed us to further improve the disambiguation accuracies we obtained. 

As discussed earlier, the first layer takes as input the root and the morphological features of a word. 
The morphological features of words we use are presented in \Tab{tab:features}. Specifically, the set of morphological features we consider contains the root, main POS tag, minor POS tag, person and possessive agreements, plurality, gender, case marker, polarity and tense. 
Note that the information contained in a surface word form may differ due to morphological characteristics of a language. 
For instance, German and French have gender feature contrary to Turkish while Turkish words have possesive agreement and polarity.
\textit{Main POS tag}, describes the category of a word and can take on values such as noun, verb, adjective and adverb. 
\textit{Minor POS tag} determines the minor morphological properties of a word such as semantic markers, causative markers and post-position. 
"Since", "While", "Propernoun", "Without" can be given as examples to this kind of morphological features in Turkish. 
\textit{Person} and \textit{possessive agreement} are used to answer the questions \qt{who} and \qt{whose} respectively, \ie they are used to indicate a person or an ownership relationship. 
\textit{Case marker} relate the nouns to the rest of the sentence as prepositions do in English. 
Nominative (none), dative(to, for), locative (at, in, on), ablative (from, out of) and genitive (of) forms are examples of the forms that can be observed in a sentence. 
\textit{Polarity} of a word is positive if the word is not negated and negative otherwise.   
\textit{Tense} indicates the tense of the verbs such as present, past and future tense. Additionally, we consider the moods of the verbs within tense feature. Moods express the speaker's attitude such as indicative, imperative or subjunctive moods. 
In languages with grammatical gender such German and French, every noun is associated with a \textit{gender}. The morphological analyzer we use associates each French word with one of the two genders (masculine and feminine) while it associates each German word with one of the four possible genders (masculine, feminine, neuter and no gender). 
Some of the suffixes in Turkish change word meaning creating derivational boundaries in the morphological analyses. 
The morphological features of a word given in \Tab{tab:features} are extracted after the final derivational boundary. 
In Turkish, we add one more feature to each word named \textit{previous tags} in order to account for the previous suffixes that the word might have. This way, our model learns the effect of suffixes that change word meaning. 
Some of the described morphological features exist only for certain word categories. For instance \textit{possessive agreement} and \textit{case marker} features can only exist in nouns, \textit{polarity} and \textit{tense} exist in verbs and \textit{person agreement} exist in nouns and verbs. If a morphological feature cannot be extracted from a word, we label it as having \textit{NULL} for the feature.

\section{Experiments}
\label{sec:Experimental Results}

For Turkish, we used a semi-automatically disambiguated corpus containing 1M tokens \cite{Yuret:2006:LMD:1220835.1220877}. 
Since this dataset is annotated semi-automatically, it also contains noise. 
In order to reduce the effect of noise to the recognition accuracies, we created a test set by randomly selecting sentences containing 20K of the tokens and manually annotating them. 
We make this test data publicly available \footnote{http://indigof:8080/Genie/disambiguationTestSet.html} so that Turkish morphological disambiguation algorithms can be compared more accurately in the future. 

We use SPMRL 2014 dataset \cite{seddah2014introducing} for German and French. This data set is created in the Penn tree bank format and used for a shared task on statistical parsing of morphologically rich languages. This dataset contains 1M and 500K sentences with POS tag and morphological information for German and French respectiveley. It provides 90\% of all sentences as training set and \%10 of rest of the sentences as test set. We align the features in the tree bank to the HFST outputs in order to determine the correct morphological analyses generated by the HFST tool. We use this data set for both training and testing.
The development sets for each language are randomly separated from the training data and are used to optimize the embedding lengths of morphological features.

We noticed that similar parameters lead to the best performance. Thus, in the experiments, we used embedding lengths 50, 20 and 5 for roots, POS tags and the other morphological features respectively. The number of filters in the first and second layers are 30 and 40 respectively. The window length, \textit{n}, that determines the number of words input to the second layer is set to 5. 

\begin{table}[!ht]
	\caption{POS tagging, lemmatization and morphological disambiguation accuracies of the proposed approach for Turkish, German and French. }
	\centering	
	\begin{tabular}{P{0.23\linewidth}ccc}
		\toprule
		 &{Turkish(\%)}&{German(\%)}&{French(\%)}\\ 
		\midrule
		{POS Tagging}                  & \postagaccuracyturkish                                                    & \postagaccuracygerman                                                    & \postagaccuracyfrench                                                     \\
		{Lemma.}                & \stemmingaccuracyturkish                                                     & \stemmingaccuracygerman                                                    & \stemmingaccuracyfrench                                                    \\
		{M. disamb.} & \accuracyturkish                                                       & \accuracygerman                                                   &  \accuracyfrench                                \\ 
		\bottomrule
	\end{tabular}
	\label{tab:results}
\end{table}

The experiment results for POS tagging, lemmatization and morphological disambiguation in Turkish, German and French are presented in \Tab{tab:results}. Notice that the POS tagging and lemmatization accuracies are refer to the percentages of POS tags and lemmas predicted correctly while morphological disambiguation accuracies are refer to the percentages of the words disambiguated correctly among the ambiguous words
According to the results, we observe that even though our initial target was to be able to achieve Turkish morphological disambiguation, our model consistently obtains high accuracies in French and German as well. 

\begin{table}[!ht]
	\caption{The comparison of the disambiguation accuracy of the proposed approach with the state-of-the-art models in Turkish. }
	\centering
	\begin{tabular}{P{0.62\linewidth}c}	
		\toprule 
		{Method} & {Accuracy(\%)} \\	
		\midrule
		{ Multilayer Perceptron}&82.13\\
		{ Decision List}&83.31\\
		{ Proposed Model - w/o pre-training}& \accuracyturkish \\
		{ Proposed Model - with pre-training}&\textbf{\accuracyWithRootEmb}\\
		\bottomrule
	\end{tabular}
	\label{tab:resultsturkish}
\end{table}	

In \Tab{tab:resultsturkish}, we present the results of various models for Turkish morphological disambiguation on our hand-labeled test data. 
The results of the multilayer perceptron developed in \cite{sak-et-al-cicling-07} and the decision list learning algorithm developed in \cite{Yuret:2006:LMD:1220835.1220877} are presented in lines 1 and 2 respectively. 
We present Turkish morphological disambiguation results obtained by our model with and without pre-training in lines 3 and 4 respectively. 
As we discussed before, unsupervised pre-training of the embeddings can boost accuracies of neural networks. As expected, morphological disambiguation accuracy increases by around 1\% (around 6\% reduction in error) when root embeddings are pre-trained instead of randomly initialized. We see that even without unsupervised pre-training our algorithm outperforms the current state of the art models and we are able to further improve the accuracy by pre-training of the embeddings. 

Although we do not evaluate the effects of unsupervised pre-training for German and French, it is expected that higher accuracies can be achieved using unsupervised pre-training of the embeddings for these languages as well.
Error analysis for Turkish morphological disambiguation shows that the root is incorrectly decided in 30\% of errors. The root is correct but the POS tag is incorrectly decided in 40\% of errors while 30\% of errors caused by wrong decisions on other inflectional groups. 
When compared with the study of Sak \etal \citeyear{sak-et-al-cicling-07}, there is no significant difference in the distribution of mistakes. However our method performs better in root decisions due to unsupervised learning of root embeddings.	
As discussed before, the available data for Turkish morphological disambiguation task contains some systematic errors. Yüret and Türe (\citeyear{Yuret:2006:LMD:1220835.1220877}) report that the accuracy of the training data is below 95\%.	
According to our observation there is a major confusion between noun and adjective POS tags in training data which affects the decision of the morphological disambiguation systems.
In our experiment, we observe that 18\% of the errors are caused by such confusion, whereas the ratio of these errors are reported as 22\% in the experiments of Sak \etal (\citeyear{sak-et-al-cicling-07}).	

\section{Summary and Future Work}
\label{sec:Conclusion}
In this paper, we present a model capable of learning word representations for languages with rich morphology. 
We show the utility of our approach in the task of Turkish, German and French morphological disambiguation. 
We also show the effect of unsupervised pre-training on recognition accuracies and improve the current state-of-the-art in Turkish morphological disambiguation. 
We publicly make available a manually annotated test set containing 20K tokens which we believe will benefit Turkish NLP. 

This paper presents a deep learning architecture specifically aiming to handle morphologically rich languages. 
Nonetheless, NLP systems that work on languages such as English can also benefit from our work. Using our model, English words can be separated into morphemes so that they can be better represented. This allows creating systems that are less affected from problems such as data sparsity \cite{DBLP:conf/conll/LuongSM13}. 

While using pre-training, we only considered the pre-trained root embeddings. It would be preferred to pre-train all the embeddings using our text corpus which we leave as future work. 
Another point of note is the selected embedding sizes that we used in our experiments. While we worked on a development set separated from training data for parameter selection, further investigation in parameter selection might improve the obtained accuracies. 

\section{ Acknowledgments}
This project is partially funded by 3140951 numbered TUBITAK-TEYDEB (The Scientific and Technological Research Council of Turkey – Technology and Innovation Funding Programs Directorate).

\bibliographystyle{aaai}
\small{
	\bibliography{main}
}

\end{document}